\documentclass[10pt,onecolumn,letterpaper]{article}
\usepackage{authblk}
\usepackage[pagenumbers]{arxiv} 

\usepackage{graphicx}
\usepackage{amsmath}
\usepackage{amssymb}
\usepackage{booktabs}
\usepackage[pdftex]{color}
\usepackage{enumerate}
\usepackage{multirow}

\usepackage[capitalize]{cleveref}
\crefname{section}{Sec.}{Secs.}
\Crefname{section}{Section}{Sections}
\Crefname{table}{Table}{Tables}
\crefname{table}{Tab.}{Tabs.}

\begin{document}

\title{Invisible-to-Visible: Privacy-Aware Human Segmentation using Airborne Ultrasound via Collaborative Learning Probabilistic U-Net}

\author[,1]{Risako Tanigawa \thanks{tanigawa.risako@jp.panasonic.com}}
\author[,1]{Yasunori Ishii \thanks{ishii.yasunori@jp.panasonic.com}}
\author[,1]{Kazuki Kozuka \thanks{kozuka.kazuki@jp.panasonic.com}}
\author[,2]{Takayoshi Yamashita \thanks{takayoshi@isc.chubu.ac.jp}}
\affil[1]{Technology Division, Panasonic Holdings Corporation}
\affil[2]{Machine Perception and Robotics Group, Chubu University}

\maketitle

\begin{abstract}
Color images are easy to understand visually and can acquire a great deal of information, such as color and texture. 
They are highly and widely used in tasks such as segmentation. 
On the other hand, in indoor person segmentation, it is necessary to collect person data considering privacy.
We propose a new task for human segmentation from invisible information, especially airborne ultrasound.
We first convert ultrasound waves to reflected ultrasound directional images (ultrasound images) to perform segmentation from invisible information.
Although ultrasound images can roughly identify a person's location, the detailed shape is ambiguous.
To address this problem, we propose a collaborative learning probabilistic U-Net that uses ultrasound and segmentation images simultaneously during training, closing the probabilistic distributions between ultrasound and segmentation images by comparing the parameters of the latent spaces.
In inference, only ultrasound images can be used to obtain segmentation results.
As a result of performance verification, the proposed method could estimate human segmentations more accurately than conventional probabilistic U-Net and other variational autoencoder models.

\end{abstract}

\section{Introduction}
Segmentation has attracted wide attention because of its applications, such as medical image diagnoses, robotics, and action recognition~\cite{shervin2020segsurvey,desouza2002robotsurvey,linlin2020robotsurvey,Intisar2020medsurvey,feng2021medsurvey,zhang2019actionsurvey}. 
A camera is a well-developed sensor for segmentation tasks.
Although camera-based segmentation has been widely investigated and achieved high precision, camera images do not preserve privacy for human segmentation.

Audio signals is a privacy-preserved segmentation~\cite{Vasudevan2020sounddepth,Irie2019sts,laput2018ubicoustics,moreira2020acoustic,sim2015acoustic,yatani2012bodyscope}.
Predicting depth maps and segmentations from audio signals have been proposed~\cite{Vasudevan2020sounddepth,Irie2019sts}.
These methods can generate images from invisible physical information and can be used to recognize human actions by analyzing segmentation images.
Although these methods can visualize sounding objects, detecting non-sounding objects are difficult.
From the human recognition standpoint, people who do not make sounds, such as not talking or walking, cannot be detected.

Airborne ultrasound echoes could be used to detect non-sounding people.
There are methods for detecting the surrounding information by analyzing ultrasound echoes~\cite{Liu2020acloc,Das2017gesture,murray2017bio,Hwang2019BatGnet,fu2015opportunities,parida2021image,xie2021hearfit}.
Although these methods can estimate the position of objects, methods that specialize in human recognition and estimate human segmentation have not been well investigated.
If human segmentation can be estimated from echoes, it would be possible to detect non-sounding people and it can be used to estimate human action.

\begin{figure}[t]
  \centering
  \includegraphics[width=0.5\linewidth]{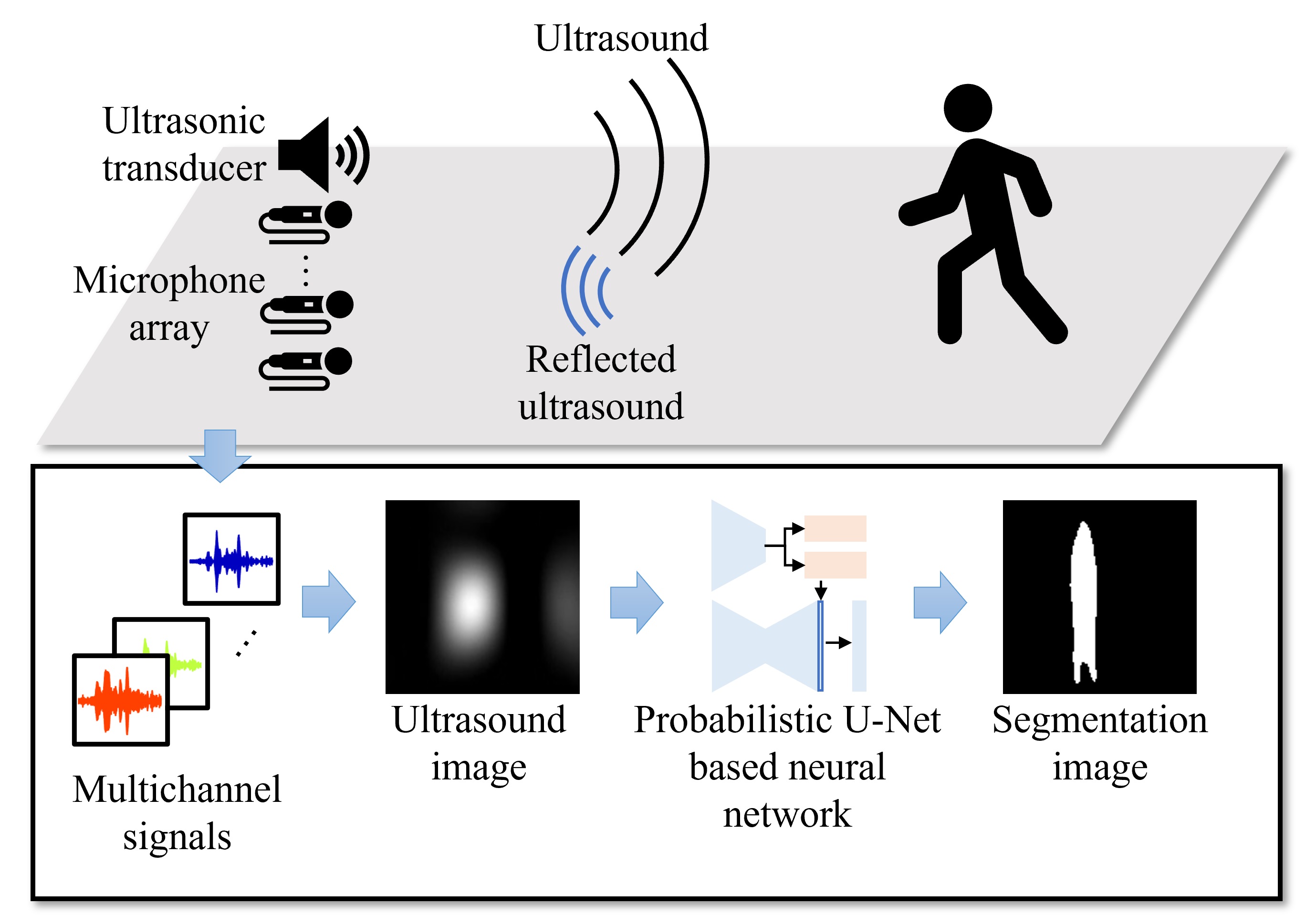}

   \caption{Concept of our work. The top row shows the setting of a sensing environment. The bottom row illustrates the concept of our method.}
   \label{fig:concept}
\end{figure}

In this study, we propose a method for estimating human segmentation from airborne ultrasound detected by multichannel microphones using a neural network.
The concept of our work is illustrated in \Cref{fig:concept}.
In the sensing section, an ultrasonic transducer at 62 kHz resonance, and 16 channels of microelectromechanical system (MEMS) microphone array are used.
The ultrasound emitted by the ultrasonic transducer is reflected by a human body and then captured by microphones.
The reflected ultrasound directional images (hereinafter, ultrasound images) can be obtained by analyzing the differences between multichannel ultrasound signals.
Because the ultrasound image represents the intensity of the reflected wave at each pixel, it has ambiguous shapes that are far from humans.
Therefore, we introduce a deep neural network to obtain human segmentation images from ultrasound images.
We use probabilistic models for the task because the ultrasound images have different edge positions from segmentation images.
Kohl et al.~\cite{Simon2018unetvae} proposed a probabilistic U-Net that combines variational autoencoder (VAE)~\cite{kingma2014vae} and U-Net~\cite{ronneberger2015unet}.
During the training phase, the probabilistic U-Net learns a network to match latent spaces of input and the input/ground truth by comparing the latent spaces with Kullback-Leibler divergence (KLD).
A latent vector, sampled from the latent space of input/ground truth, combined with the last activation map of a U-Net.
The spatial positions of the ultrasound and segmentation images are similar and the edges are different.
Hence, the latent distributions of ultrasound and segmentation images are roughly close because of shrinking the spatial dimensions by convolutions.
Thus, reducing the distance between latent distributions with precisely estimating the edges is difficult.
Therefore, we propose a collaborative learning probabilistic U-Net (CLPU-Net), which uses mean squared error (MSE)  instead of KLD to minimize the distance between latent distributions.
Experiments showed that human segmentation images could be generated from ultrasound images.
To the best of our knowledge, this is the first work to estimate human segmentation images from airborne ultrasounds.

The main contributions to this work are as follows:
\begin{itemize}
      \item A human body sensing system using an airborne ultrasound and a microphone array for privacy-aware human segmentation.
      \item An architecture of CLPU-Net that learns a network with matching latent variables from segmentation and ultrasound images.
      \item We showed that despite a simple mean-squared error, the distance between latent variables can be shortened by expressing the distribution of high-dimensional latent space with mean and variance.
\end{itemize}

\section{Related work}


\subsection{Privacy-preserved human segmentation}
The camera-based methods have been well investigated and have achieved high accuracy. However, privacy concerns should be considered for applications such as home surveillance.
Cameraless human segmentation methods have also been investigated to address privacy issues.
Wang et al.~\cite{wang2019personinwifi} proposed a method for estimating human segmentation images, joint heatmaps, and part affinity fields using Wi-Fi signals.
They used three transmitting-receiving antenna pairs and thirty electromagnetic frequencies with five sequential samples.
The channel state information~\cite{Halperin2011csi} was analyzed to input networks.
The networks comprise upsampling blocks, residual convolutional blocks, U-Net, and downsampling blocks.
Although this method achieved privacy-friendly fine-grained person perception with Wi-Fi antennas and routers, the Wi-Fi signals are highly affected by the surrounding environment due to the multipath effect.

Alonso et al.~\cite{alonso2018evsegnet} proposed an event camera-based semantic segmentation method.
The event camera detects pixel information when the brightness changes, such as when the subject moves.
They do not capture personal information more clearly than the RGB cameras because event cameras only capture changes in intensities on a pixel-by-pixel basis.
In~\cite{alonso2018evsegnet}, event information from event cameras was formed as $6$-channel images.
The first two channels were histograms of positive and negative events, whereas the other four channels were the mean and standard deviation of normalized timestamps at each pixel for the positive and negative events.
They showed that an Xception-based encoder--decoder architecture could learn semantic segmentation from the $6$-channel information.
Although this method achieves semantic segmentation from privacy-preserved event cameras, it is difficult to detect people who are not moving.

Irie et al.~\cite{Irie2019sts} proposed a method that generates segmentation images from sounds.
They recorded sounds using $4$-channel microphone arrays.
They used Mel frequency cepstral coefficients and angular spectrum from sounds for estimating segmentation images.
This method can estimate human and environmental objects only from sounds.
However, estimating segmentation images for non-sounding people is difficult in principle because this method analyzes the sound emitted from objects.

\subsection{Airborne ultrasonic sensing}
Airborne ultrasound could be used to detect non-sounding people.
The positions of people can be detected by analyzing ultrasound echos reflected by them.
Although the multipath effect influences detection, it is less effective than radio waves because the propagation speed of ultrasound is slower than that of radio waves.

Airborne ultrasonic sensors have been used to detect distance in various industries, such as the automobile~\cite{Wang2014parking} and manufacturing industries~\cite{FANG2017uswood,CHIMENTI2014usmaterial,KAZYS2006usmultilayer}.
Ultrasonic sensors emit short pulses at regular intervals.
The ultrasounds are reflected if there are objects in the propagation path.
The distance of objects can be determined by analyzing the time differences between emitted and reflected sounds~\cite{Carullo2001US}.

When sounds are captured using microphone arrays, the directions of objects can be detected in addition to their distances~\cite{Moebus20073dus}.
Sound localization using beamforming algorithms has been developed. 
A delay-and-sum (DAS) method is a common beamforming algorithm~\cite{Perrot2021das}.
The DAS method can estimate the direction of sound sources by adding array microphone signals delayed by a given amount of time.
By using ultrasonic transducers and microphone arrays, the positions of objects can also be estimated using the DAS with regarding a reflected position as a sound source.
Although this method can detect the position of objects, it is difficult to obtain the actual shape of objects from echoes of a single pulse.

Hwang et al.~\cite{Hwang2019BatGnet} performed three-dimensional shape detection using wideband ultrasound and neural networks.
Although the analysis of multiple frequencies can precisely detect the positions and shapes of objects, the sound system for emitting wideband frequency becomes large, and the number of data increases because of the high sampling rate required to sense wideband ultrasound.
Therefore, we consider obtaining segmentation images from the positional information on reflected objects analyzed by narrowband frequency ultrasound using a neural network.

\section{Ultrasound sensing in the proposed method}
First, we describe the hardware setup of our ultrasound sensing system.
Then, we describe the preprocessing for converting ultrasound waves to ultrasound images.

\subsection{Ultrasound sensing system}
\begin{figure}[t]
  \centering
  \includegraphics[width=0.5\linewidth]{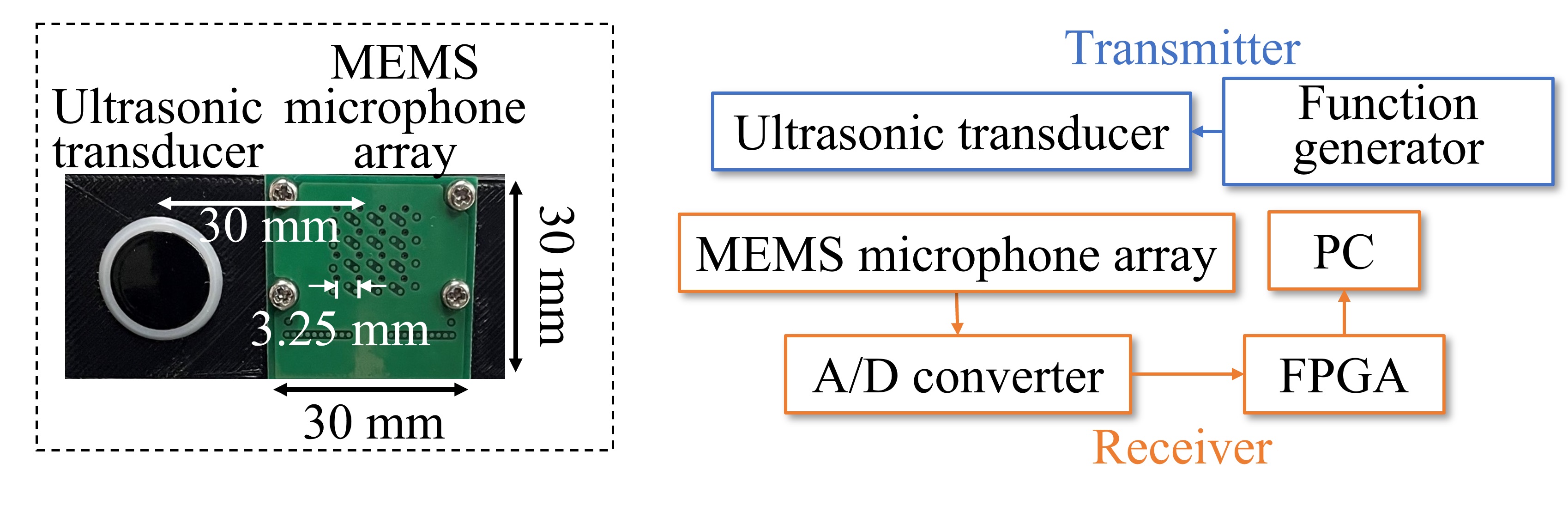}
   \caption{Hardware setup of ultrasonic wave transmitter and receiver system.}
   \label{fig:us_sensor}
\end{figure}

The hardware setup is shown in \Cref{fig:us_sensor}.
The transmitter comprised a function generator and an ultrasonic transducer at 62 kHz resonance.
The ultrasonic transducer was driven with burst waves of $20$ cycles and $50 \, \rm{ms}$ intervals at $62 \, \rm{kHz}$ using the function generator.
The receiver comprised a MEMS microphone array, an analog-to-digital converter, a field-programmable gate array (FPGA), and a PC.
A $4 \times 4$ grid MEMS microphone array, whose microphones were mounted on a $30 \, \rm{mm^2}$ substrate at $3.25 \, \rm{mm}$ intervals, was used.
The analog signals captured by the microphones were converted to digital signals and imported into the PC through the FPGA.
The distance between the microphone array and ultrasonic transducer was set to $30 \, \rm{mm}$.

\begin{figure*}[t]
  \centering
  \includegraphics[width=1.0\linewidth]{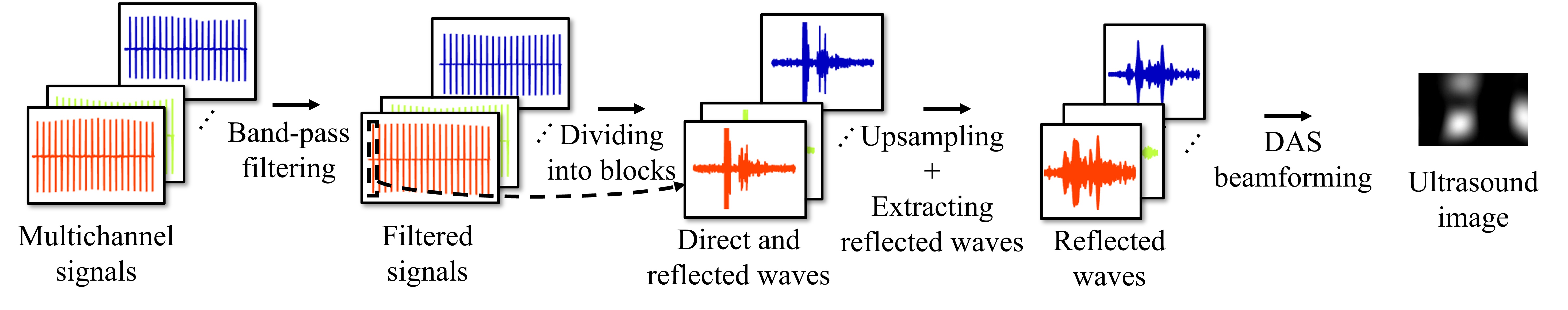}

   \caption{Diagram of preprocessing of ultrasound signals. The multiple ultrasound signals are filtered and then divided into single direct-reflected wave pairs. The upsampled reflected waves are extracted and DAS beamformed, and an ultrasound image is created.}
   \label{fig:preprocess_arch}
\end{figure*}

\subsection{Data preprocessing}
The diagram of the data preprocessing is shown in \Cref{fig:preprocess_arch}.
First, a band-pass filter with a center frequency of $62 \, \rm{kHz}$ and bandwidth of $10 \, \rm{kHz}$ was used for signals captured by the $16$ microphones.
The filtered ultrasound signals were divided into blocks including direct and reflected waves.
Then, we upsampled the ultrasound waves four times at each block to improve the accuracy of direction estimation.
Following that, we produced ultrasound images from reflected ultrasounds via DAS beamforming.
The beamformed signal $y$ can be defined as
\begin{equation}
	y(t) = \sum_{m=1}^{M} x_m(t-\Delta_m),
\end{equation}
where $t$ is the time, $M$ is the number of microphones, $x_m$ is the signal received by the $m$-th microphone, and $\Delta_m$ is the time delay for the $m$-th microphone, which is determined by the speed of sound and distance between the $m$-th microphone and observing points.
We set the observing points in the range of $\theta=-45$--$45$ degrees in the azimuthal direction and $\phi=-60$--$60$ degrees in the polar direction.
Then, we calculated beamformed signals and obtained the reflected directional heat maps.
To reduce the noise from reflected waves from objects other than people, we calculated ultrasound heat maps $H_{\rm{us}}$ by subtracting a reference map $H_{\rm{ref}}$ from reflected directional heat maps $H$ as
\begin{equation}
  H_{\rm{us}}(i,j) = H(i,j) - k H_{\rm{ref}}(i,j),
\end{equation}
where $(i,j)$ is the pixel of the heat maps, and $k$ is the coefficient, which is determined by
\begin{equation}
	k = \frac{H(i_{\rm{max}}, j_{\rm{max}})}{H_{\rm{ref}}(i_{\rm{max}}, j_{\rm{max}})},
\end{equation}
where $(i_{\rm{max}}, j_{\rm{max}})$ is the index of the maximum pixel of $H_{\rm{ref}}$.
Notably, the reference map was calculated using the data without people, and $H_{\rm{us}}$ was normalized as
\begin{equation}
	X_{\rm{us}}(i, j) =
	\begin{cases}
		0, & H_{\rm{us}}(i,j) < 0 \\
		\frac{H_{\rm{us}}(i,j)}{\max(H_{\rm{us}})}, & H_{\rm{us}}(i,j) \geq 0
	\end{cases}
  \label{eq:soundimage}
\end{equation}
when it was converted to ultrasound images $X_{\rm{us}}$.

\begin{figure}[t]
  \centering
  \includegraphics[width=0.5\linewidth]{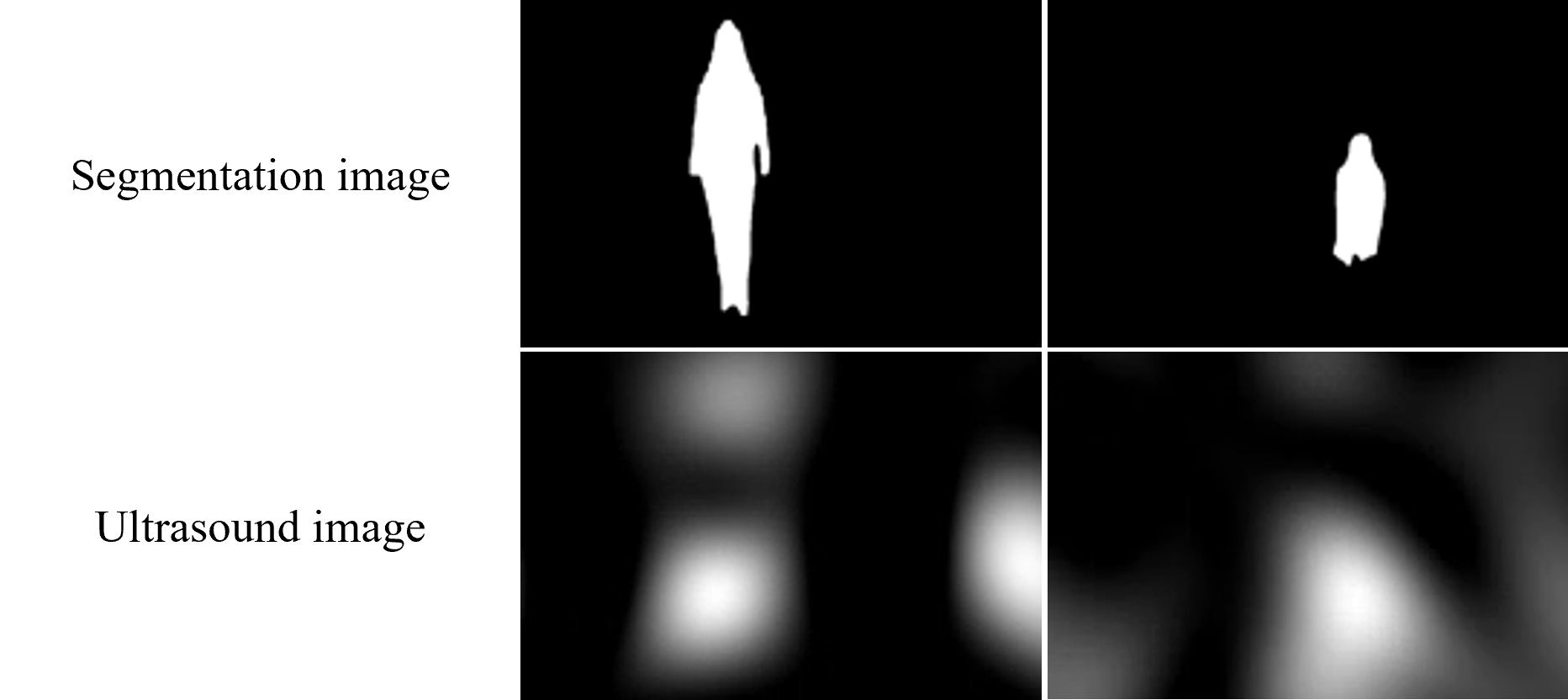}

   \caption{Example of segmentation and ultrasound images. The top row is the segmentation images genera by RGB images and the bottom row is the ultrasound images.}
   \label{fig:soundimage}
\end{figure}

Examples of segmentation and ultrasound images are shown in \Cref{fig:soundimage}.
The segmentation images in the top row were annotated from RGB images.
The ultrasound images, which represent the intensity of the reflected ultrasound at each pixel, had ambiguous shapes at the positions corresponding to a person.
Furthermore, there were artifacts in the region without any people.

\section{Human segmentation via ultrasound}
We first describe an overview of the proposed network.
Second, we briefly explain probabilistic U-Net, which is the basis of the proposed method.
Finally, we describe the proposed network in detail.

\subsection{Overview}
As shown in \cref{fig:soundimage}, the ultrasound images do not have the shape of a person, and their edges are very different from those of the segmentation images.
When there is a lot of ambiguity in the input image or when the difference between input and output is large, it is better to use the probabilistic method than the deterministic method.
Thus, we consider performing human segmentation based on probabilistic models.
Kohl et al.~\cite{Simon2018unetvae} proposed a probabilistic U-Net combining VAE~\cite{kingma2014vae} with U-Net.
This method learns prior and posterior networks, which output the latent spaces of input and input/segmentation images, respectively, to be close.
The latent variables sampled from the latent space of the posterior network are added to the last layer of U-Net. 
This method deals with large discrepancies between the input and output images by sampling latent variables from the latent space output by the prior network during the inference phase.
However, it is difficult to handle with the existing prior network because the difference in appearance between the ultrasound and segmentation images is huge.
Thus, we propose collaborative learning probabilistic U-net based on probabilistic U-Net.
Details of the proposed method are described below.

\begin{figure}[t]
  \centering
  \includegraphics[width=1.0\linewidth]{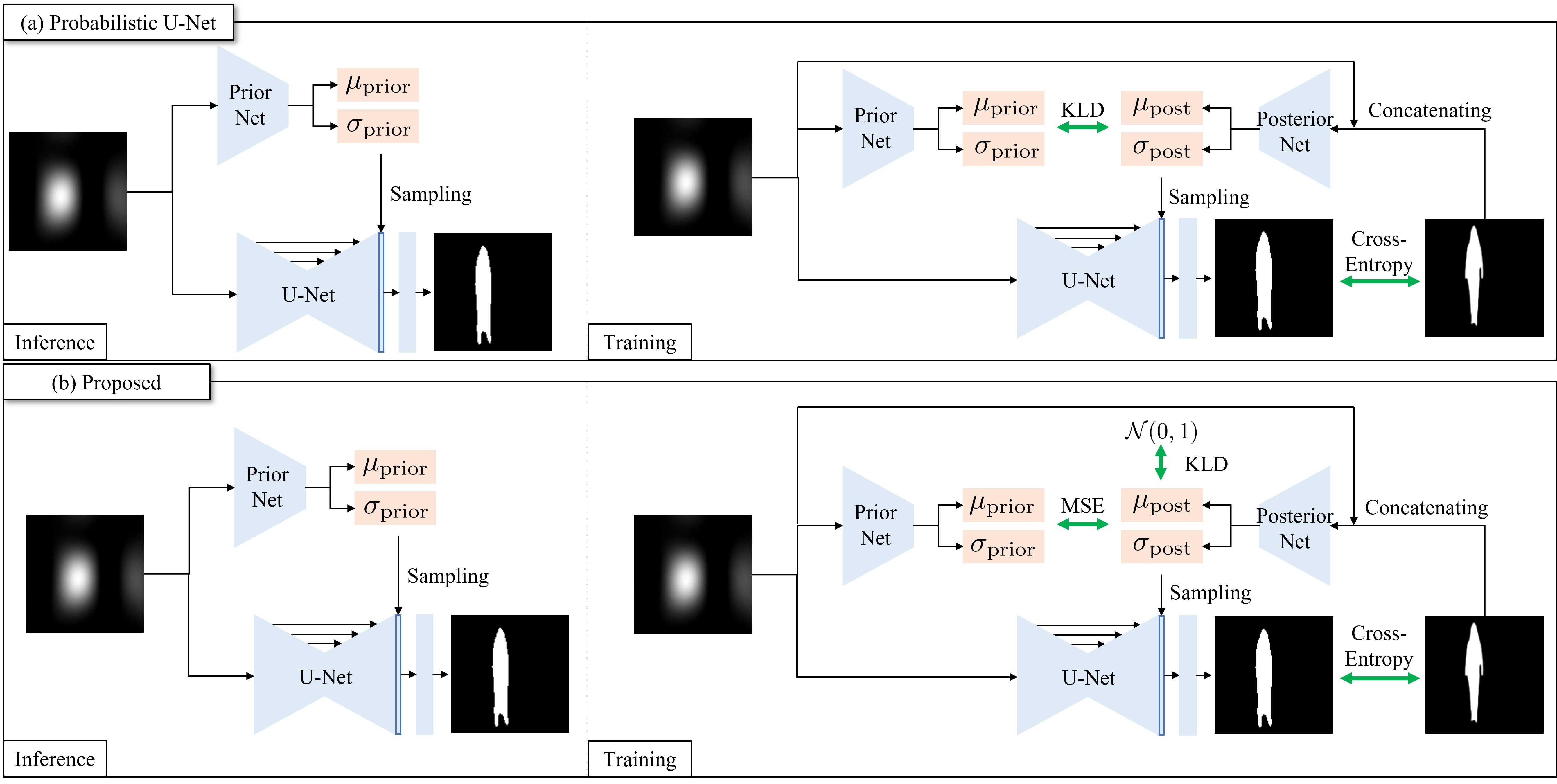}

   \caption{Network architecture. (a) is the conventional probabilistic U-Net and (b) is the proposed CLPU-Net.}
   \label{fig:network}
\end{figure}

\subsection{Probabilistic U-Net~\cite{Simon2018unetvae}}
In this section, we explain the probabilistic U-Net.
The architecture of the probabilistic U-Net is shown in \Cref{fig:network}(a).
During the inference phase, the input image $X_{\rm{in}}$ is input into the prior network and U-Net.
The prior network, which is parameterized by $\omega$, encodes the input image and outputs the parameters of the mean $\mu_{\rm{prior}}(X_{\rm{in}}; \omega)$ and variance $\sigma_{\rm{prior}}(X_{\rm{in}}; \omega)$.
Then the data $z$ was randomly sampled from the prior distribution $P$ as
\begin{equation}
	z \sim P(\cdot | X_{\rm{in}}) = \mathcal{N} \bigl( \mu_{\rm{prior}}(X_{\rm{in}}; \omega),\rm{diag}(\sigma_{\rm{prior}}(X_{\rm{in}}; \omega)) \bigr),
\end{equation}
and broadcasted to the N-channel feature map to match the shape to the segmentation image.
The feature maps are concatenated at the last layer of the U-Net, and the output image is estimated by convoluting the feature maps from the prior network and U-Net.

During the training phase, the posterior network is added to encode the information about the segmentation image.
The posterior network parameterized by $\nu$ encodes the data that contains the input image $X_{\rm{in}}$ and segmentation image $X_{\rm{seg}}$.
The parameters of the mean $\mu_{\rm{post}}(X_{\rm{in}}, X_{\rm{seg}}; \nu)$ and variance $\sigma_{\rm{post}}(X_{\rm{in}}, X_{\rm{seg}}; \nu)$ of the posterior distribution $Q$ are obtained using the posterior network.
Then, the data $z$ is sampled from the posterior distribution $Q$ as
\begin{equation}
	z \sim Q(\cdot | X_{\rm{in}}, X_{\rm{seg}}) = \mathcal{N} \bigl( \mu_{\rm{post}}(X_{\rm{in}}, X_{\rm{seg}}; \nu),\rm{diag}(\sigma_{\rm{post}}(X_{\rm{in}}, X_{\rm{seg}}; \nu)) \bigr),
\end{equation}
and combined to the feature map at the last layer of the U-Net.
The loss function consists of two terms. The first term is cross entropy to minimize the error between the estimated image and ground truth. The second term is KLD to close the distribution between the prior distribution $P$ and posterior distribution $Q$.
The losses are combined as a weighted sum,
\begin{equation}
	L(X_{\rm{seg}}, X_{\rm{in}}) = \mathbb E_{z \sim Q(\cdot | X_{\rm{seg}}, X_{\rm{in}})} \left[ - \log P(X_{\rm{seg}}|X_{\rm{out}}(X_{\rm{in}}, z)) \right] \nonumber \\
+ \beta D_{\rm{KL}} (Q(z|X_{\rm{seg}}, X_{\rm{in}}) \parallel P(z|X_{\rm{in}})),
\end{equation}
where $X_{\rm{out}}(\cdot)$ outputs the estimated segmentation image and $\beta$ is the weight parameter.

\subsection{CLPU-Net}


Although the appearance of ultrasound and ground truth segmentation images differ significantly at the edges, the appearance of the other parts, particularly positions, is relatively similar.
Therefore, it is important to learn latent space by focusing on the difference in edges.
In probabilistic U-Net, prior distribution $P$ and posterior distribution $Q$ are penalized using KLD.
Reducing the distance between the distributions to estimate edges with high accuracy by comparing the distributions obtained after the spatial dimensions decreased at the prior/posterior network.
Because the ultrasound and segmentation images are roughly matched other than edges.
Thus, we propose a method to use mean squared error (MSE) of means and variances instead of KLD.

\begin{figure}[t]
  \centering
  \includegraphics[width=1.0\linewidth]{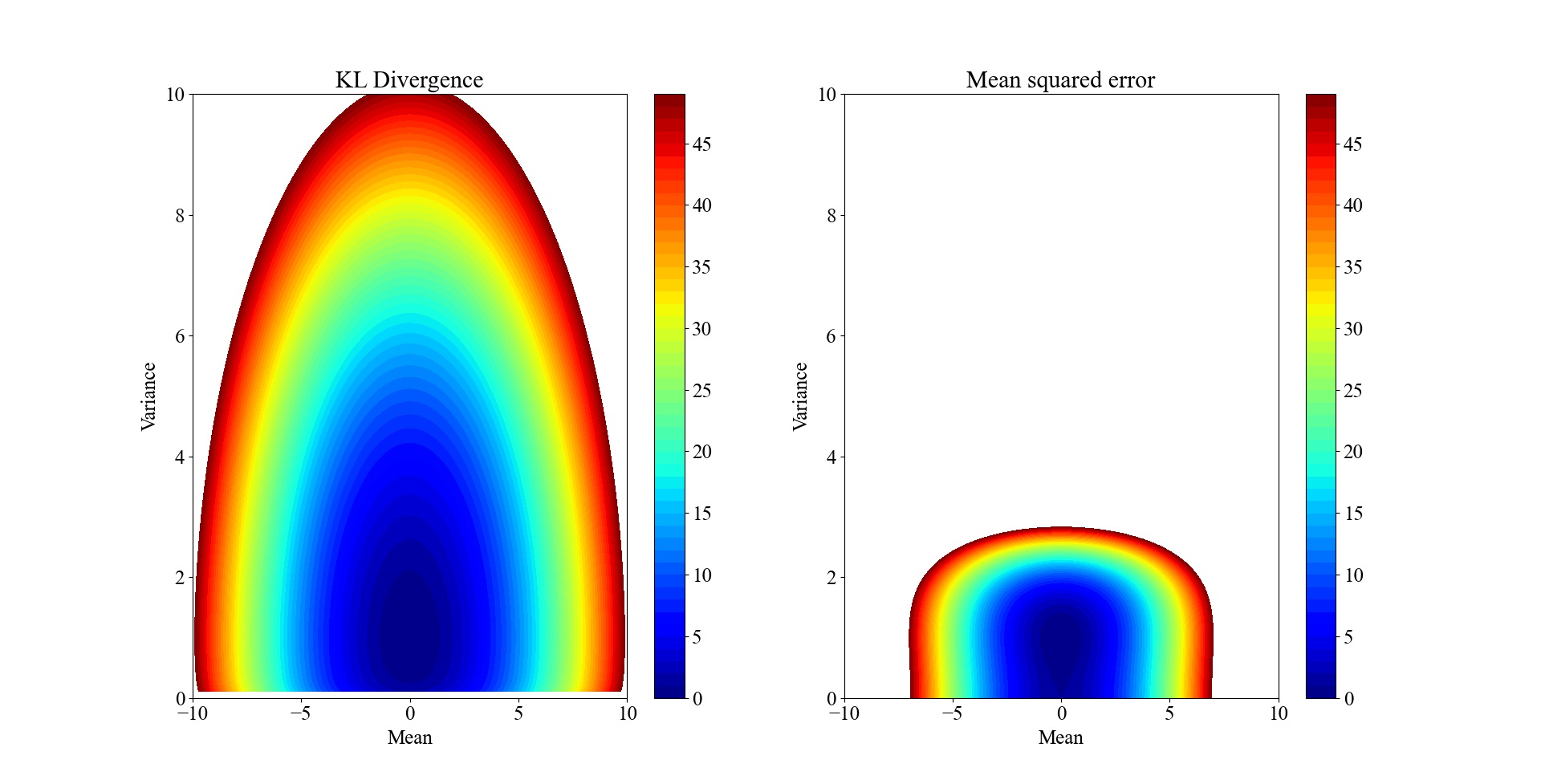}

   \caption{Comparison of errors with standard normal distribution by KLD and MSE. The left figure shows the errors calculated by KLD, and the right figure shows the errors calculated by MSE.}
   \label{fig:kld-mse}
\end{figure}

The errors between the standard normal distribution calculated by KLD and MSE are shown in \Cref{fig:kld-mse}.
The MSE loss was calculated as
\begin{equation}
	\rm{MSE} = (\mu - \mu_0)^2 + (\sigma - \sigma_0)^2,
\end{equation}
where $\mu_0 = 0$ and $\sigma_0 = 1$ are the mean and variance of the standard normal distribution.
Since the value around the error of 0 for the MSE loss changes more rapidly than that of the KLD (\cref{fig:kld-mse}), the MSE loss is more sensitive than the KLD.
Therefore, converging to an optimal result by KLD is difficult because of the small gradient of loss landscape, since the ultrasound and segmentation images are roughly matched other than edges.
Thus, we propose a method that uses MSE loss of the means and variances to penalize the differences of the latent spaces.
Additionally, KLD regularization is added to the posterior distribution to approach a standard normal distribution.

The proposed network is illustrated in \Cref{fig:network}(b).
The comparison method between the prior and posterior distributions at the training phase distinguishes the probabilistic U-Net.
The loss function $L$ of the proposed method is as follows:
\begin{equation}
	L = \alpha L_{\rm{VAE}} + (1-\alpha) L_{\rm{MSE}},
\end{equation}
where $\alpha$ is the weights adjusting the scale.
The first term $L_{\rm{VAE}}$ is
\begin{equation}
	L_{\rm{VAE}} = \mathbb E_{z \sim Q(\cdot | X_{\rm{seg}}, X_{\rm{us}})} \left[ - \log P(X_{\rm{seg}}|X_{\rm{out}}(X_{\rm{us}}, z)) \right] \nonumber \\
+ \beta D_{\rm{KL}} (P_0(z) \parallel Q(z|X_{\rm{seg}}, X_{\rm{us}})),
\end{equation}
where $X_{\rm{us}}$ is the ultrasound image and $P_0(z)$ is the standard normal distribution.
The second term $L_{\rm{MSE}}$ is 
\begin{equation}
	L_{\rm{MSE}} = \frac{1}{N} \sum_{n=1}^{N} (\mu_{\rm{prior}\it{,n}} - \mu_{\rm{post}\it{,n}})^2 + \frac{1}{N} \sum_{n=1}^{N} (\sigma_{\rm{prior}\it{,n}} - \sigma_{\rm{post}\it{,n}})^2,
\end{equation}
where $N$ is the dimension of the latent vector.


\section{Experiments}
In this section, we describe the dataset configuration, experimental setup, and evaluation results.

\begin{table}[t]
  \centering
  \caption{Number of images at each condition.}
  \label{tab:numImgs}
  \begin{tabular}{c|c|c|c|c|c|c|c}
   \noalign{\smallskip} \hline \noalign{\smallskip}
   Condition number & 1 & 2 & 3 & 4 & 5 & 6 & total \\ \noalign{\smallskip} \hline \noalign{\smallskip}
   \, Number of images \, & \, 7,792 \, & \, 7,768 \, & \, 7,634 \, & \, 7,727 \, & \, 7,777 \, & \, 7,796 \, & \, 46,494 \, \\ \noalign{\smallskip}
   \hline
  \end{tabular}
\end{table}

\subsection{Experimental setup}
\subsubsection{Datasets}
\begin{figure}[t]
  \centering
  \includegraphics[width=0.9\linewidth]{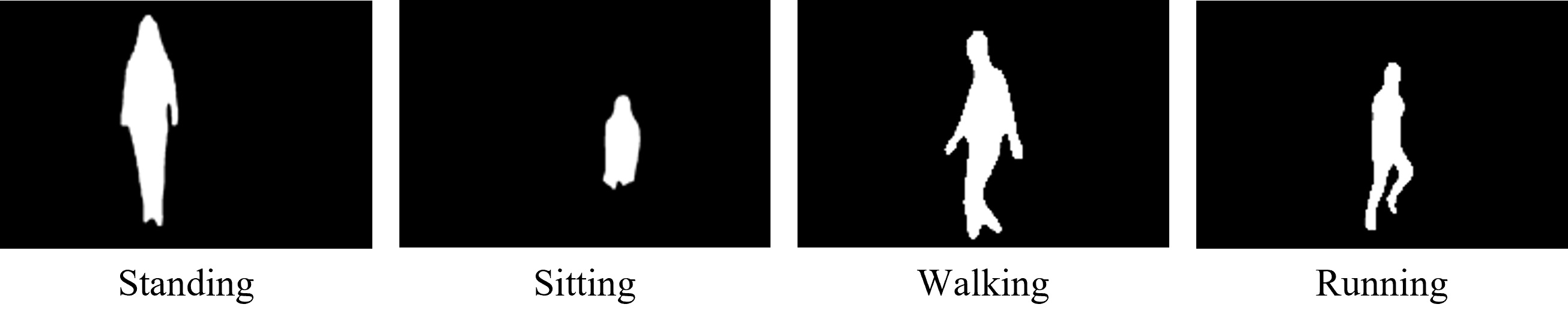}
   \caption{Segmentation images. The motions are standing, sitting, walking, and running in order from the left.}
   \label{fig:img_example}
\end{figure}

\begin{figure}[t]
  \centering
  \includegraphics[width=0.9\linewidth]{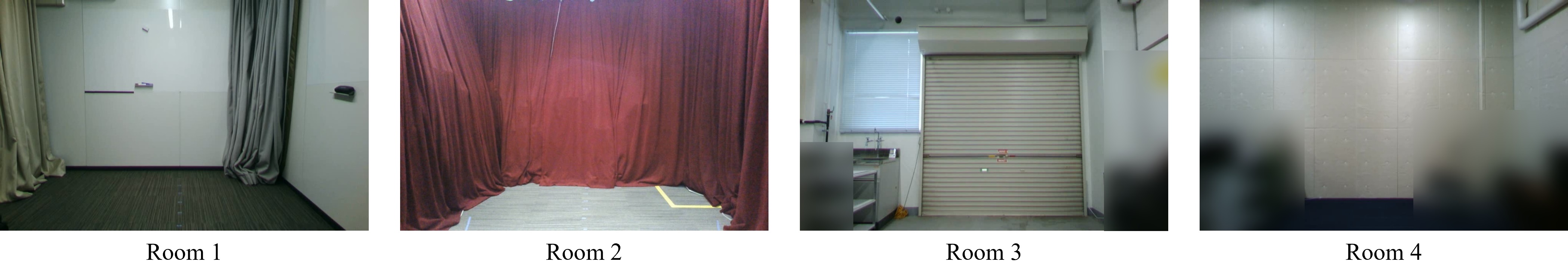}
   \caption{Rooms used for data acquisition. Confidential areas are blurred.}
   \label{fig:room}
\end{figure}

We created a dataset\footnote{
This data acquisition experiment was judged by our institution to be not the subject of the examination by the Institutional Review Board.
Participants in data acquisition have given their written consent.
} because no datasets have been previously used airborne ultrasound to detect humans.
For $10 \, \rm{s}$, we captured the ultrasounds at $192 \, \rm{kHz}$ sampling from $16$ channel microphones and videos at $30$ frames per second (fps) from the RGB camera (a built-in camera of Let's Note, CF-SV7, Panasonic), which was located $35 \, \rm{mm}$ under the microphone array.
The resolution was $180 \times 120$, and the videos were used for creating segmentation images for the training phase.
The data were extracted at $10$ fps because the time interval of the ultrasound generation was $20$ bursts per second and the frame rate of the video was $30$ fps.
We automatically produced segmentation images using Mask R-CNN~\cite{he2018mask}.
We used the dataset that people, who were located from $1$ to $3 \, \rm{m}$ away from the sensing devices, performed motions such as standing, sitting, walking, and running.
Examples of the segmentation images were shown in Figure~\ref{fig:img_example}.
These are the images for standing, sitting, walking, running, respectively.
There were six participants and they performed in four different rooms.
The rooms are shown in \Cref{fig:room}.
Rooms 1 and 3 were surrounded by relatively acoustically reflective walls and rooms 2 and 4 were surrounded by relatively acoustically absorbent materials.
The number of images at each condition is shown in \Cref{tab:numImgs}.
The condition numbers $1$ to $6$ indicate the participants' number.
Images were captured about 7,700 images at each condition, the total number was 46,494.

\begin{table}[t]
  \centering
  \caption{IoU, accuracy, precision, recall, F1-socre of VAE, Joint-VAE, Probabilistic U-Net, and CLPU-Net. Each metric was averaged for all six patterns of the k-fold cross-validation.}
  \label{tab:IoU}
  \begin{tabular}{c|ccccc}
   \hline \noalign{\smallskip}
   Model & IoU & \, Accuracy \, & \, Precision \, & \, Recall \, & \, F1-score \, \\ \noalign{\smallskip} \hline \noalign{\smallskip}
   VAE & \, 0.265 \, & 0.880 & 0.351 & 0.526 & 0.403 \\ \noalign{\smallskip}
   Joint-VAE & 0.278 & 0.889 & 0.376 & 0.507 & 0.392 \\ \noalign{\smallskip}
   \, Probabilistic U-Net \, & 0.329 & 0.912 & 0.485 & 0.490 & 0.456 \\ \noalign{\smallskip}
   CLPU-Net & 0.388 & 0.921 & 0.536 & 0.546 & 0.519 \\ \noalign{\smallskip}
   \hline
  \end{tabular}
\end{table}

\subsubsection{Evaluation}
We evaluated the proposed method using k-fold cross-validation.
To confirm the robustness of the unknown person data, the dataset was divided based on the person, which corresponds to the conditions listed in \cref{tab:numImgs}.
Therefore, $k$ was set to six, and all six patterns were trained and evaluated, for example learning at condition number 2 to 6 and evaluating at condition number 1.
The performance of the model was evaluated using an intersection-over-union (IoU), accuracy, precision, recall, and F1-score.

\subsubsection{Implementation details}
The prior and posterior networks consisted of four blocks of layers.
A block consisted of three pairs of a 2-dimensional convolution layer and ReLU and an average pooling layer.
The output channels of the four blocks were $32, 64, 128, 192$ in order.
All input images were resized to $128 \times 128$ pixels.
The dimension of means $\mu_{\rm{prior}}, \mu_{\rm{post}}$ and variances $\sigma_{\rm{prior}}, \sigma_{\rm{post}}$ was set to $20$.
The batch size was $16$ and the initial learning rate was $0.0001$.
We used an Adam optimizer with $\beta_1 = 0.9 , \beta_2 = 0.999$ in training.
$\alpha$ was $0.0001$ and $\beta$ was $0.3$, respectively, which are determined by a grid search.

\subsection{Experimental results}
We first describe the performance of the proposed CLPU-Net and compare it with other methods.
Then, we described the comparison of the accuracy by distances between the sensor and person.
Then, we explain the comparison of the accuracy by rooms.
Finally, quantitative results are described.

\subsubsection{Performance of CLPU-Net}
We first evaluated the performance of the proposed CLPU-Net.
Then, we compared our model with probabilistic U-Net, VAE, Joint-VAE~\cite{dupont2018jointVAE} to confirm the validity of learning with ultrasound and segmentation images.
The VAE and Joint-VAE were trained using segmentation images and inferred using ultrasound images.
The CLPU-Net and probabilistic U-Net were trained using ultrasound and segmentation images and inferred using ultrasound images.
\Cref{tab:IoU} illustrates the qualitative results.
Each metric was averaged for all six patterns of the k-fold cross-validation.
The proposed CLPU-Net was marked highest in all metrics of the four models.

\begin{figure}[t]
  \centering
  \includegraphics[width=1.0\linewidth]{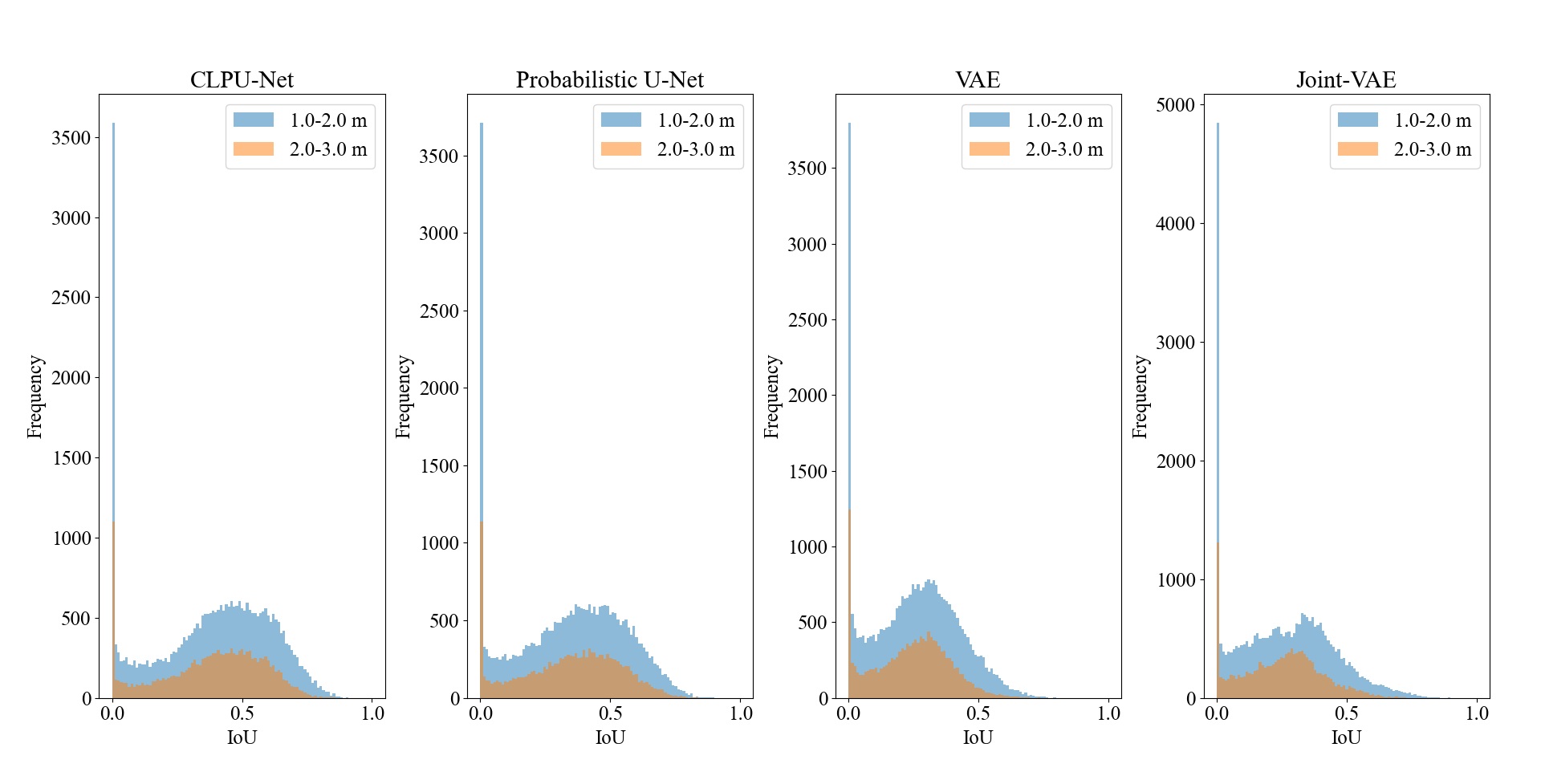}
   \caption{Histograms of IoU of CLPU-Net and conventional methods. The frequency distributions of the data with distances between $1$ m and less than $2$ m, and the data of distances between $2$ m and less than $3$ m were displayed.}
   \label{fig:hist_analysis}
\end{figure}


\subsubsection{Comparison by distances between sensor and person}
The amplitude of the reflected ultrasound decreases as the distance between the sensors and person increases.
Therefore, we investigated the robustness of the distance between the sensor and a person.
\Cref{fig:hist_analysis} shows the IoU histogram, which categorizes the dataset into two types; distances between $1$ m and less than $2$ m and distances between $2$ m and less than $3$ m.
The horizontal axis represents the IoU values, and the vertical axis represents the frequency, which means the number of images at each bin.
The bin width was set to $0.01$.
The blue bins were the data of distances between $1$ m and less than $2$ m, and the orange bins were the data of distances between $2$ m and less than $3$ m.
Although the frequencies differ due to the difference in the number of images for the conditions, the trends in these frequency distributions were not very different.
Therefore, the accuracy is not affected by the distance within a range of $1$ to $3$ m.
We verified that robustness is sufficient in the range of use because we assume an application will be used at home.

\begin{figure}[t]
  \centering
  \includegraphics[width=1.0\linewidth]{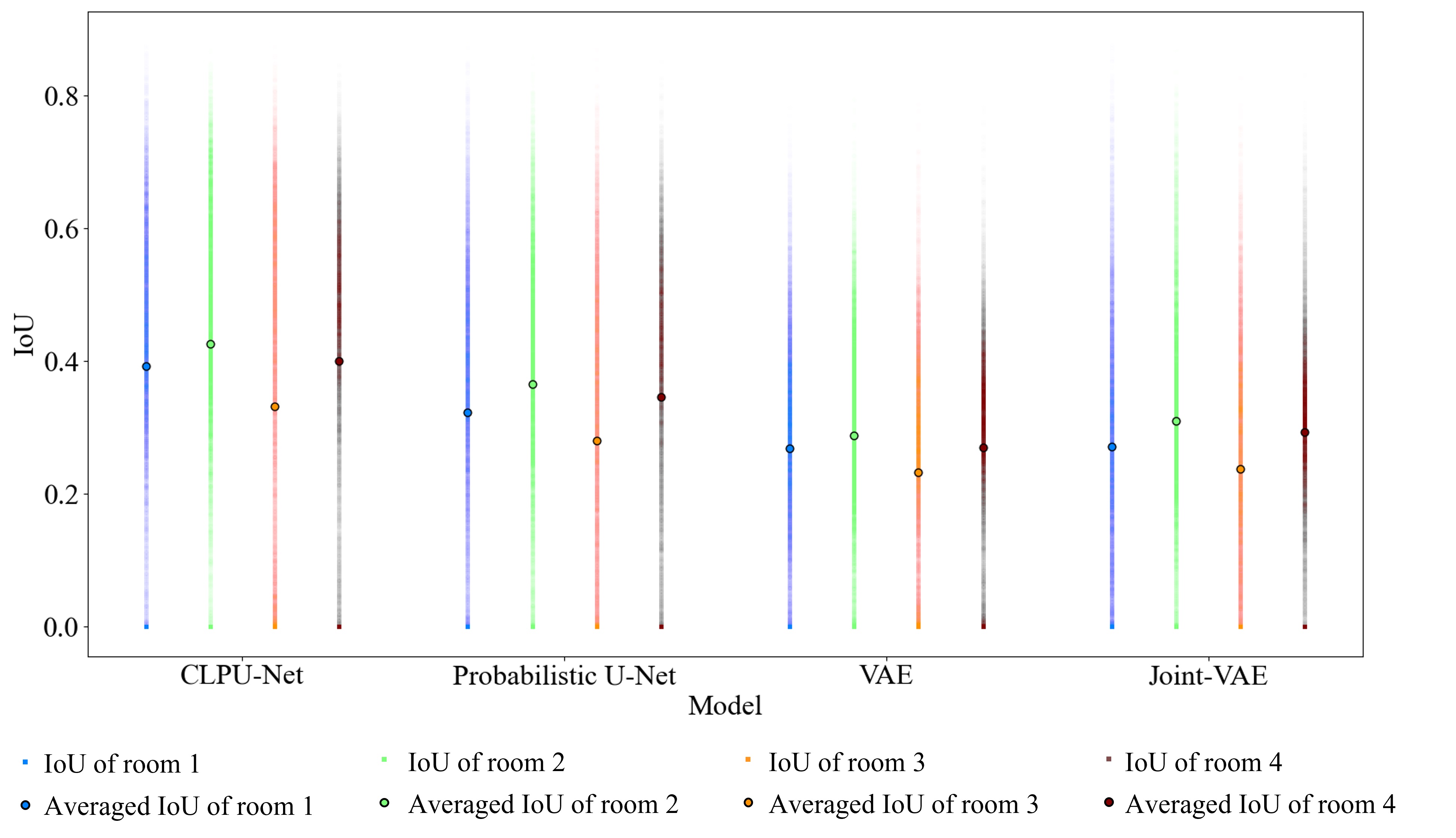}
   \caption{IoUs of rooms $1$, $2$, $3$, and $4$ of CLPU-Net and conventional methods. The semi-transparent square represents the IoU of an estimated image, and the circle with a black line represents the averaged IoU. The blue, green, orange, and brown represent the IoUs of rooms $1$, $2$, $3$, and $4$, respectively.}
   \label{fig:site_analysis}
\end{figure}

\begin{figure}[t]
  \centering
  \includegraphics[width=1.0\linewidth]{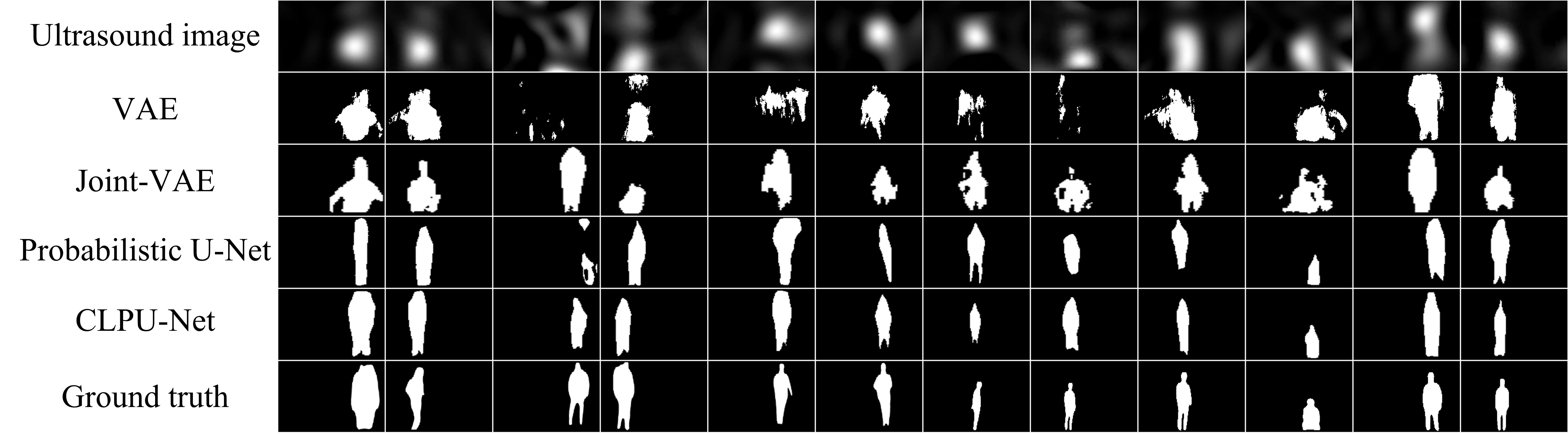}
   \caption{Quantitative results. The top row is ultrasound images, the second to fourth rows are estimated images by conventional methods, the fifth row is estimated images by CLPU-Net, and the bottom row is ground truth.}
   \label{fig:est_imgs}
\end{figure}

\subsubsection{Comparison by rooms}
There are various acoustic absorption characteristics in rooms assuming an application will be used at home.
Therefore, we investigated the robustness against the room differences.
The IoU for each room is illustrated in Figure~\ref{fig:site_analysis}.
The IoU of all estimated images were plotted as semi-transparent squares, and the averaged IoU was plotted as a circle with a black line.
The blue, green, orange, and brown represent the IoU of rooms $1$, $2$, $3$, and $4$, respectively. 
The accuracies of rooms $2$ and $4$, which have a relatively high acoustic absorption rate, tend to be higher and the accuracies of rooms $1$ and $3$, which have a relatively low acoustic absorption rate, tend to be lower.
The decrease in accuracy in reverberant rooms seems to be the effect of higher-order reflected and environmental ultrasounds.
The accuracies can be improved in reverberant rooms by changing the analysis section of ultrasound waves at preprocessing according to the distance between the sensor and person.

\subsubsection{Quantitative result}
The quantitative result is shown in \Cref{fig:est_imgs}.
In the VAE and Joint-VAE, the shapes are not properly estimated and there are no people in some estimated images.
In these methods, the information on ultrasound images was excluded at the training phase, and they cannot estimate segmentation images from ultrasound images.
Therefore, using the information in the segmentation images at the training phase affects the estimation.
Although the probabilistic U-Net can capture the shapes better than those of the VAE and Joint-VAE, 
the estimation fails if the input and segmentation images have a large discrepancy.
Alternatively, the proposed method has been estimated under such conditions.
Some images from CLPU-Net, such as $1$st, $5$th, $8$th, and $12$th columns from the right, are closer to the ground truth.
In contrast, those images from probabilistic U-Net tend to swell or shrink.

\section{Conclusions}
We proposed privacy-aware human segmentation from airborne ultrasound using CLPU-Net.
Our method used the MSE of the means and variances, which are the output of the prior and posterior networks, instead of KLD for comparison of the latent spaces of the prior and posterior networks in probabilistic U-Net.
This enables optimization suitable for the ultrasound image obtained by our proposed device.
This method can be used to detect human actions 
in situations where privacy is required, such as home surveillance because the sound/segmentation images cannot be reconstructed to RGB images.
In the future, we will increase various data and improve the ultrasound image generation process to improve the accuracy in unknown environments.
{\small
\bibliographystyle{ieee_fullname}
\bibliography{egbib}
}

\end{document}